\documentclass[10pt,twocolumn,letterpaper]{article}

\usepackage{pgfplotstable}
\usepackage{iccv}
\usepackage{times}
\usepackage{epsfig}
\usepackage{graphicx}
\usepackage{amsmath}
\usepackage{amssymb}
\usepackage{xspace}
\usepackage{multirow}
\usepackage{subfig}
\usepackage{onimage}
\usepackage{tikz}
\usepackage{pgfplots} 
\usetikzlibrary{positioning,calc}
\usepgfplotslibrary{colorbrewer}

\newcommand{\sgen}{$\mathcal{S}_{\text{src}}$\xspace}
\newcommand{\stask}{$\mathcal{S}_{\text{task}}$\xspace}

\usepackage[breaklinks=true,bookmarks=false]{hyperref}

\iccvfinalcopy 

\ificcvfinal\fi

\begin{document}

\title{Learning Effective Visual Relationship Detector on 1 GPU}

\author{
Yichao Lu~\thanks{Authors contributed equally and order is determined randomly.}\\
Layer6 AI\\
{\tt\small yichao@layer6.ai}
\and
Cheng Chang~\footnotemark[1]\\
Layer6 AI\\
{\tt\small jason@layer6.ai}
\and
Himanshu Rai~\footnotemark[1]\\
Layer6 AI\\
{\tt\small himanshu@layer6.ai}
\and
\newline{}
Guangwei Yu\\
Layer6 AI\\
{\tt\small guang@layer6.ai}
\and
Maksims Volkovs\\
Layer6 AI\\
{\tt\small maks@layer6.ai}
}

\maketitle
\ificcvfinal\thispagestyle{empty}\fi

\begin{abstract}
We present our winning solution to the Open Images 2019 Visual Relationship
challenge. This is the largest challenge of its kind to date with nearly 9 million
training images. Challenge task consists of detecting objects and identifying
relationships between them in complex scenes. Our solution has 
three stages, first object detection model is fine-tuned for the challenge classes using 
a novel weight transfer approach. Then, spatio-semantic and visual relationship models 
are trained on candidate object
pairs. Finally, features and model predictions are combined to generate the final
relationship prediction. Throughout the challenge we focused on minimizing the
hardware requirements of our architecture. Specifically, our weight transfer approach 
enables much faster optimization, allowing the entire architecture to be trained 
on a single GPU in under two days. In addition to efficient optimization, our approach 
also achieves superior accuracy winning first place out of over 200 teams, 
and outperforming the second place team by over $5\%$ on the held-out private leaderboard.

\end{abstract}

\section{Introduction}

Visual relationship detection is a core computer vision task that has gained a
lot of attention recently~\cite{chen2019scene,yang2018graph,zellers2018scenegraphs,zhang2019vrd}. 
The task comprises of object detection followed by
visual relationship prediction to identify relationships between pairs of objects.
Relationship identification involves inferring complex spatial, semantic and visual 
information between objects in a given scene, which is a challenging task. 
Successfully solving this task is a natural first step towards scene understanding 
and reasoning. The Open Images 2019 Visual Relationship challenge introduces a uniquely 
large and diverse dataset of annotated images designed to benchmark visual relationship
models in a standardised setting. The challenge dataset is based on the Open Images
V5 dataset~\cite{kuznetsova2018open}, which contains 9 million images annotated with 
class labels, bounding boxes, segmentation masks and visual relationships.

The challenge task is to detect objects and their associated
relationships. The relationships include human-object relationships (e.g. ``man
holding camera''), object-object relationships (e.g. ``spoon on table''), and
object-attribute relationships (e.g. ``handbag is made of leather'').
Each of the relationships can be expressed as a triplet, written as a pair of
objects connected by a relationship predicate e.g. (``beer", ``on", ``table"). Visual
attributes are also triplets where object is connected with an attribute
using the ``is'' relationship  e.g. (``table", ``is", ``wooden''). The challenge 
contains 329 unique triplets, which span 57 different object classes, 5
attributes, and 10 predicates. In this paper we present our solution which ranked 
first out of over 200 teams, and outperformed the second place team by over $5\%$ on 
the held-out private leaderboard. To make our approach more practical, we focus
on minimizing the hardware requirements during training. Specifically, we show that 
through transfer learning we can significantly speed up optimization, and train the
entire model on a {\it single} GPU in under two days.
\begin{figure*}
    \centering
    \includegraphics[width=0.9\textwidth]{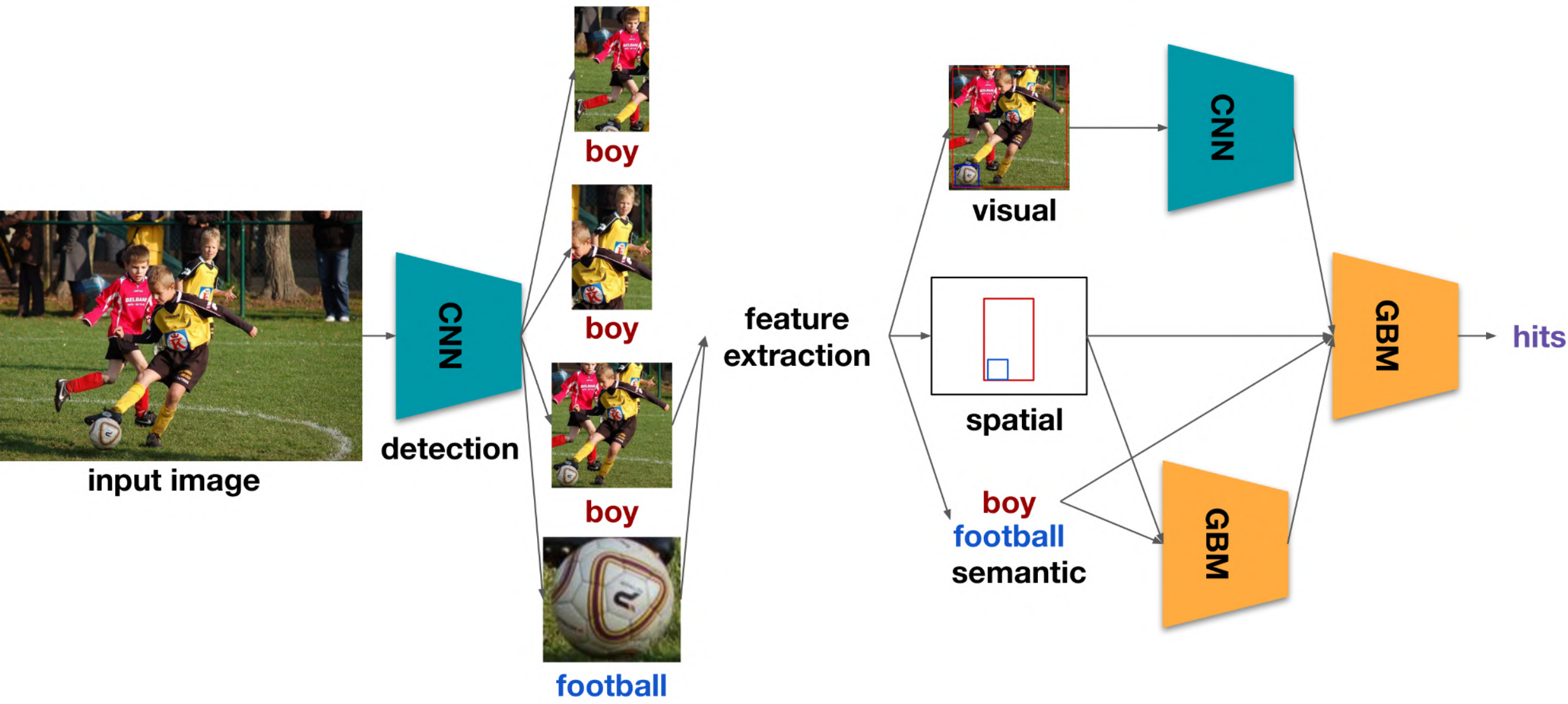}
  \caption{Proposed visual relationship model architecture. Object detection 
  model is first applied to get bounding boxes for every object in the input image. Spatio-semantic and
  visual models are then applied to candidate object pairs to generate initial relationship predictions. 
  Finally, features and model predictions are combined in the last stage to output the final relationship 
  prediction. In this example our model outputs (``boy", ``hits", ``football").}
  \label{fig:diagram}
\end{figure*}


\section{Our Approach}

Our pipeline consists of three stages. In the first stage, object detection model 
fine-tuned for the challenge classes generates object bounding boxes along with their associated confidences. 
In the second stage, two separate models based on gradient boosting and convolutional 
neural networks are used to model spatio-semantic and visual features. Finally, a third stage 
model takes outputs from the first two stages as input and generates the final prediction. 
In this section, we describe each stage in detail, and Figure \ref{fig:diagram} summarizes 
the entire pipeline.

\subsection{Object Detection With Partial Weight Transfer}\label{sc:pwt}

Training object detection model from scratch in a reasonable amount of time on a large
dataset requires significant resources. For instance, one of the leading 
models on the Open Images 2018 Object Detection Track needed 33 hours of
training on 512 GPUs~\cite{akiba2018pfdet}. Instead of running multiple 
costly and time consuming training experiments, we focus speeding up
optimization with limited hardware resources. In order to achieve this, 
we propose the partial weight transfer strategy. The main idea is to transfer 
as much information from models trained on well-established object detection 
benchmarks such as COCO~\cite{DBLP:journals/corr/LinMBHPRDZ14} 
and Open Images~\cite{kuznetsova2018open}. Minimal fine-tuning is then 
performed on the target task dataset.

Transfer learning is a popular and economic approach for improving
generalization by transferring knowledge between datasets and domains. 
However, choosing which model parts to keep or discard can have a 
significant impact on the performance. A common approach in object detection
is to use a popular model such as Faster RCNN~\cite{renNIPS15fasterrcnn}
or Retina Net~\cite{focalRetina} pre-trained on large and general datasets where 
high performance is observed. Then the classification and regression heads are 
replaced with randomly initialized weights to train task-specific detectors. 
However, we observed that fine-tuning the network this way can still take significant 
amount of time to converge, and doesn't always achieve high accuracy. To improve 
convergence we propose to also initialise classification and regression heads with 
pre-trained weights by approximately matching classes between datasets.

\begin{figure}[t]\centering
    \includegraphics[width=0.4\textwidth]{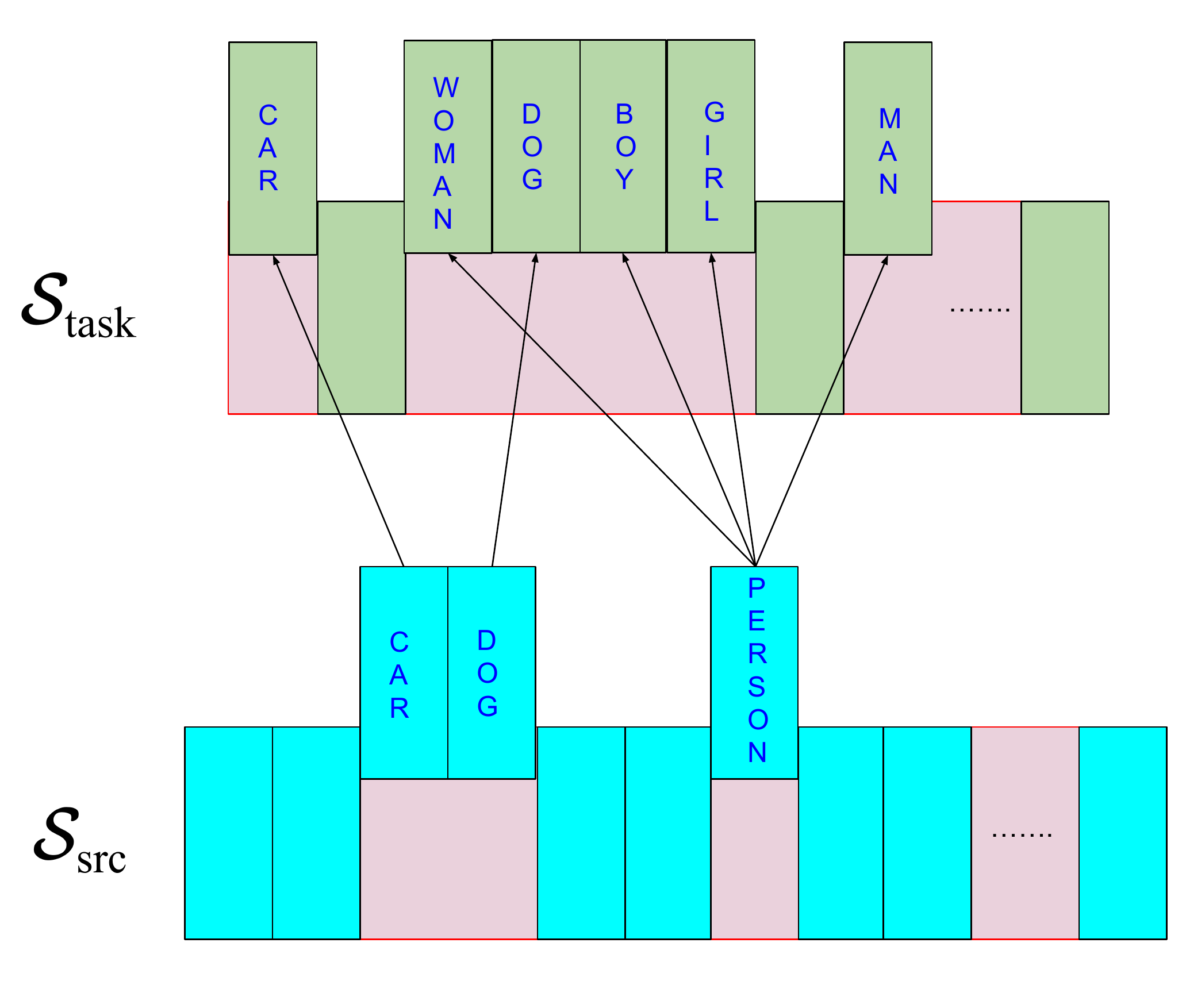}
  \caption{Partial weight transfer diagram example. Classification head weights for
  ``car" and ``dog" classes are directly copied to corresponding classes in \stask.
  Weights for the more general class ``person" are extrapolated to related 
  classes ``woman", ``boy", ``girl", and ``man" in \stask.}
   \label{fig:diag}
\end{figure}

We denote source and target task datasets as \sgen and \stask respectively. 
\sgen is typically a large public dataset such as COCO on which many of
the leading models are trained and released. \stask in our case is the
challenge dataset, and the aim is to transfer models from \sgen 
to \stask with high accuracy and minimal computational resources. The
main difficulty here is that the target dataset typically 
contains classes not present in \sgen. However, we hypothesize that 
there should be common information learned by the model for 
related classes between the two datasets. Following this intuition, it 
should be possible to partially transfer model weights for 
related classes from \sgen to \stask and improve fine-tuning.
Figure~\ref{fig:diag} demonstrates this process. In this example, weight 
vectors in the classification head of \sgen model associated with 
``car" and ``dog" classes are directly copied to corresponding classes 
in \stask. Similarly, weights for the more general class ``person"
are transferred to related classes ``woman", ``boy", ``girl", and ``man" 
in \stask. Weights for classes that don't have a match are randomly initialized.
Formally, given a mapping $k \to g(k)$ from task class index $k$ to source class index, 
the classification head layer for the task model has the following structure:
\begin{align}
  z_k &= \begin{cases}
  \omega_k^\text{task} \cdot x^n + b_k^{\text{task}} & \text{if} \ \ g(k) = \varnothing\\
   \omega_{g(k)}^\text{src} \cdot x^n + b_{g(k)}^{\text{src}} &  \text{otherwise}
 \end{cases}
\end{align}
where $x^n$ is the input from previous layer, $\omega_{g(k)}^\text{src}$ and $b_{g(k)}^{\text{src}}$ 
are weights and biases transferred from the source dataset, and $\omega_k^\text{task}$ and 
$b_k^{\text{task}}$ are randomly initialized parameters. Applying this weight transfer even for 
approximate class matches such as ``person" $\to$ ``boy", enable us to 
significantly accelerate fine-tuning to less than one day on a single GPU and achieve better accuracy.
In the challenge we use detection models pre-trained on the popular COCO dataset which has 80 classes.
After matching classes between datasets we were able to transfer classification weights for 44 
of the 57 challenge classes.

Fine-tuned detection model can be independently evaluated on the related Open Images 2019 Object Detection 
challenge by submitting predictions only for classes that are common to both challenges. 
Table~\ref{table:1} 
shows detection mAP performance for the Cascade RCNN~\cite{cai18cascadercnn} model pre-trained on COCO
and fine-tuned for visual relationship classes. From the table we see that partial weight transfer significantly 
improves leaderboard performance by over 35\%. We also found that training time was considerably
reduced from around one week to less than a day. Blending multiple models with test time image augmentation
provides additional performance boost, and we use this approach as the first stage in our pipeline.

\subsection{Spatio-Semantic and Visual Models}\label{sc:spatio-semantic}

Given the bounding boxes predicted by the object detection model, the relationship model aims to
(1) detect whether a two objects are related, and (2) predict their relationship. These tasks
require simultaneously learning spatial, semantic and visual features. In our experiments, we
find tree-based gradient boosting models (GBMs) to be effective for learning spatial and 
semantic features, while convolutional neural network (CNN) models excel at capturing visual features.
The second stage in our pipeline thus utilizes both GBM and CNN models to perform feature extraction
for pairs of objects.
\begin{figure*}
    \centering
    \includegraphics[scale=0.48]{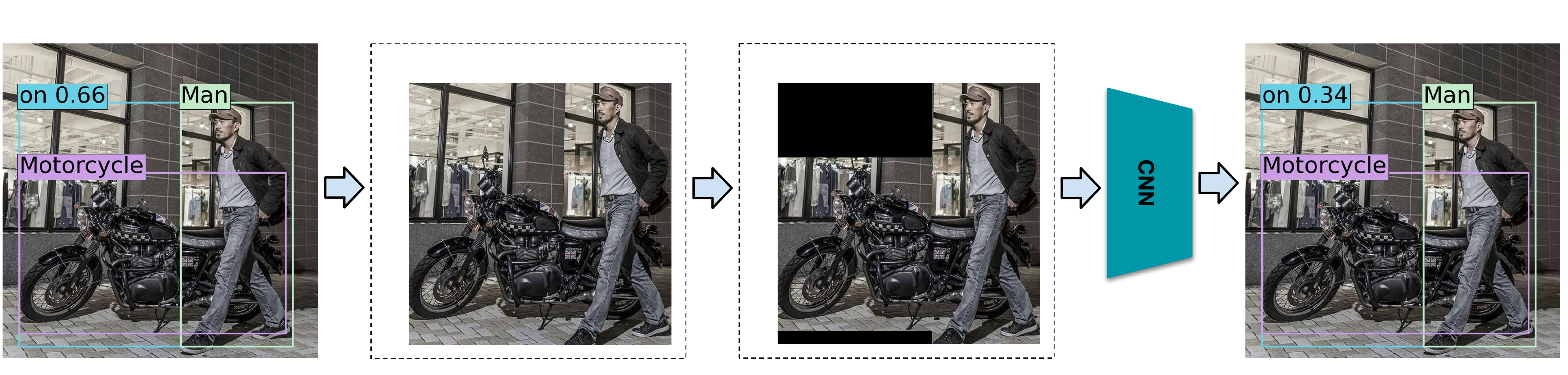}
  \caption{Visual model pipeline example on one the challenge images. Here, spatio-semantic model is 
  unable to correctly predict the relationship and outputs high probability for ``on". 
  Cropping the bounding boxes for ``man" and ``motorcycle", blacking-out the background and
  passing the resulting image through the visual CNN model reduces the probability for
  ``on" from 0.66 to 0.34.}
  \label{fig:example}
\end{figure*}

{\bf Spatio-Semantic Model.} Spatial information such as location of the object in the 
image and relative position between objects, plays an important role in relationship detection. 
Objects that are far away from each other less likely to have a relationship, and relative 
position between objects can be very informative when determining relationships such as 
``on'' or ``under''. Semantic information on the other hand, can capture the likelihood of the 
two objects co-occurring together or having a certain type of relationship. We describe both 
types of information through features and train a GBM model to predict relationship type.
The features include:
\begin{enumerate}
\item Object spatial features -  we use features such as relative and absolute position of the object 
in the image and size of the object (estimated by its bounding box).
\item Object semantic features - we include features such as other objects that this object 
typically appears with and types of relationships that in commonly has.
\item Pairwise spatial features - we encode information such as relative position of the two objects, 
IOU between their bounding boxes and Euclidean distance between box centers.
\item Pairwise semantic features - similar to pairwise spatial features, we summarize how frequently
the two objects appear together and types of relationships they typically have.
\end{enumerate}
\begin{table}[t]
\centering
\begin{tabular}{ |l||c|c|  }
 \hline
 \multicolumn{3}{|c|}{Object Detection Results (mAP)} \\
 \hline
 Model & Validation & LB\\
 \hline
 Cascade RCNN~\cite{cai18cascadercnn}     & 0.43 & 0.048\\
 Cascade RCNN~\cite{cai18cascadercnn} + PWT & 0.53 & 0.065\\ 
 Blend + TTA  & {\bf 0.56} &  {\bf 0.068}\\
 \hline
\end{tabular}
\vskip 0.1cm
\caption{Results on the Open Images 2019 Object Detection challenge. Only 57 classes from
  the visual relationship challenge are submitted. PWT and TTA denote partial weight transfer and
  test time image augmentation respectively.}
\label{table:1}
\end{table}

We consider two alternatives for defining the GBM training objective. The first option is to train a single 
GBM for multi-class classification over all the possible relationships with an additional ``None'' class 
for no relationship. The second option is to train separate GBM models for every relationship type with a 
binary classification objective. For example, for the relationship type ``under'', we find all pairs of 
objects that can possibly form an ``under'' relationship. Then label those that are present in the ground 
truth set as positive samples and others as negative samples. The advantage of the first option is that 
it is more computationally efficient and only requires a single model for all types of relationships. 
However, empirically we find that the second option performs consistently better. We presume that this 
is due to the fact it allows the model to separately focus on each relationship improving generalisation. 

{\bf Visual Model.} Models that rely solely on spatial and semantic features have failure
modes that can only be corrected with visual information. One example of such failure mode is 
shown in Figure~\ref{fig:example}. Here, both spatial and semantic features indicate that the likely
relationship is (``man", ``on", ``motorcycle"). However, from visual inspection it is clear that the 
man is actually {\it next} to the motorcycle and not on it. To incorporate visual information
we use a CNN-based architecture. Given a pair of objects for which we aim to predict the
relationship, we first crop the image so it only contains the union of the bounding boxes
for the two objects. Then for each pixel in the cropped image that does not belong to the bounding 
box of either object, we turn it into background by making it black. This reduces background 
and scene clutter, and enables the model to focus on the target objects. 
Analogous to the spatio-semantic model, we also find that training separate models (fine-tuned from
the same backbone) for each relationship type yields better results than single multi-class model.
From Figure~\ref{fig:example} we see that the visual model reduces the probability of 
(``man", ``on", ``motorcycle") from 0.66 to 0.34.

\subsection{Model Aggregation}

The last stage takes predictions from the spatio-semantic and visual models, and combines them
to make the final prediction. A straightforward way to combine models is through averaging. However,
depending on the properties of the input scene, different types of model tend to perform better and 
need to be selected accordingly. Averaging prevents such specialisation, so in the last stage we 
train another model that takes as input predictions from the second stage together with image and 
target object pair features, and learns how to optimally combine them. We also use GBM here as 
decision trees can learn highly non-smooth decision boundaries that are 
beneficial for specialisation. To train the model we split the official training set into two parts. 
All second stage models are trained on the first part, and the ensemble model is trained on the second part.

\section{Experiments}

The challenge is based on the Open Images V5 dataset which is a
large-scale multi-modal collection of over 9 million images. A subset of 
1,743,042 images contain bounding boxes, and we use this subset to train and validate 
the object detection model. The challenge dataset is a subset of the Open Images V5 data, 
and contains 391,073 labelled relationship triplets from 100,521 images. There are a total
of 329 unique triplets with 287 object-object relationships 
over 57 unique object classes, and 42 object-attribute ``is'' relationships over 5 attributes. 
Furthermore, an additional 99,999 images are used as the held-out test set that is
split 30\%/70\% for public and private leaderboards respectively. All test labels
are hidden and model evaluation is done by submitting predictions to the
Kaggle platform. Public leaderboard score is available throughout the 
competition, while private leaderboard is released at the end and used to
compute final team rankings.

\begin{figure}
    \centering
    \usepgfplotslibrary{colorbrewer}

\pgfplotsset{compat=1.11,
    /pgfplots/ybar legend/.style={
    /pgfplots/legend image code/.code={%
       \draw[##1,/tikz/.cd,yshift=-0.25em]
        (0cm,0cm) rectangle (3pt,0.8em);},
   },
}

\def\alpha{0.4}
\def\ms{0.25pt}

\begin{tikzpicture}[font=\Large,scale=0.7]
\definecolor{color0}{rgb}{0.333333333333333,0.52156862745098,0.941176470588235}
\definecolor{color1}{rgb}{0.913725490196078,0.490196078431373,0.490196078431373}

\begin{axis}[
cycle list/Dark2,
legend cell align={left},
xmin=-2, xmax=62,
ymin=-0.005,ymax=0.15,
xtick={0, 20,  40, 60},
ytick={0, 0.05,  0.1, 0.15},
ymajorgrids=true,
xmajorgrids=true,
grid style=dashed,
ylabel={$p(k)$},
xlabel={Class index $k$},
yticklabel style={
        /pgf/number format/fixed,
        /pgf/number format/precision=2,
        /pgf/number format/fixed zerofill,
},
scaled y ticks=false,
label style={font=\Large},
no markers,
legend style={nodes={scale=0.8, transform shape}},
]


\addplot+[mark=*, line width=1.5, mark size=\ms] coordinates {
(0,0.01982632142)(1,0.01982632142)(2,0.01982632142)(3,0.01982632142)(4,0.01982632142)(5,0.01982632142)(6,0.01982632142)(7,0.01982632142)(8,0.01982632142)(9,0.01982632142)(10,0.01982632142)(11,0.01982632142)(12,0.01982632142)(13,0.01982632142)(14,0.01982632142)(15,0.01982632142)(16,0.01982632142)(17,0.01982632142)(18,0.01982632142)(19,0.01982632142)(20,0.01982632142)(21,0.01982632142)(22,0.01982632142)(23,0.01982632142)(24,0.01982632142)(25,0.01982632142)(26,0.01982632142)(27,0.01982632142)(28,0.01982632142)(29,0.01982632142)(30,0.01982632142)(31,0.01982632142)(32,0.01982632142)(33,0.01982632142)(34,0.01982632142)(35,0.01982632142)(36,0.01982632142)(37,0.01982632142)(38,0.01982632142)(39,0.01982632142)(40,0.01982632142)(41,0.01982632142)(42,0.0195289266)(43,0.01871604742)(44,0.01685237321)(45,0.01683254689)(46,0.01262936675)(47,0.0124905825)(48,0.01223284032)(49,0.0108251715)(50,0.009952813355)(51,0.009615765891)(52,0.007177128356)(53,0.005828938499)(54,0.005828938499)(55,0.00557119632)(56,0.003211864071)
};
\addlegendentry{$N=1,000$}

\addplot+[mark=*, line width=1.5, mark size=\ms] coordinates {
(0,0.0357641)(1,0.0357641)(2,0.0357641)(3,0.0357641)(4,0.0357641)(5,0.0357641)(6,0.0357641)(7,0.0357641)(8,0.0357641)(9,0.0357641)(10,0.0357641)(11,0.0357641)(12,0.0357641)(13,0.0357641)(14,0.0357641)(15,0.0357641)(16,0.0357641)(17,0.0357641)(18,0.03420836165)(19,0.0276098852)(20,0.02585386789)(21,0.02290332964)(22,0.02276384965)(23,0.01905153607)(24,0.01900504274)(25,0.01822896177)(26,0.01501734559)(27,0.01274274883)(28,0.01253531705)(29,0.01170201352)(30,0.01082221666)(31,0.01011408748)(32,0.009277207539)(33,0.008923142949)(34,0.008125603519)(35,0.006112084689)(36,0.006033403669)(37,0.005479060119)(38,0.005368191409)(39,0.00492829298)(40,0.00491398734)(41,0.00434891456)(42,0.00352276385)(43,0.00337613104)(44,0.0030399485)(45,0.00303637209)(46,0.00227817317)(47,0.0022531383)(48,0.00220664497)(49,0.00195271986)(50,0.00179535782)(51,0.00173455885)(52,0.00129466042)(53,0.00105146454)(54,0.00105146454)(55,0.00100497121)(56,0.0005793784199)
};
\addlegendentry{$N=10,000$}

\addplot+[mark=*,line width=1.5, mark size=\ms] coordinates {
(0,0.05091774137)(1,0.05091774137)(2,0.05091774137)(3,0.05091774137)(4,0.05091774137)(5,0.05091774137)(6,0.05091774137)(7,0.05091774137)(8,0.05091774137)(9,0.05091774137)(10,0.05091774137)(11,0.05091774137)(12,0.05054706021)(13,0.03092336269)(14,0.0272552486)(15,0.02722673467)(16,0.02634280268)(17,0.02381524599)(18,0.01948112785)(19,0.01572339854)(20,0.01472337409)(21,0.01304308863)(22,0.01296365695)(23,0.01084955233)(24,0.01082307511)(25,0.01038110911)(26,0.008552143841)(27,0.0072567965)(28,0.00713866734)(29,0.006664113991)(30,0.006163083415)(31,0.005759814904)(32,0.005283224845)(33,0.005081590589)(34,0.004627404336)(35,0.0034807368)(36,0.003435929188)(37,0.003120239191)(38,0.003057101192)(39,0.002806585904)(40,0.002798439066)(41,0.00247663894)(42,0.00200615901)(43,0.001922653914)(44,0.001731203207)(45,0.001729166497)(46,0.00129738405)(47,0.001283127083)(48,0.001256649857)(49,0.001112043472)(50,0.001022428247)(51,0.0009878041826)(52,0.000737288895)(53,0.0005987926385)(54,0.0005987926385)(55,0.000572315413)(56,0.0003299469641)
};
\addlegendentry{$N=25,000$}

\addplot+[mark=*,line width=1.5, mark size=\ms] coordinates {
(0,0.4311744127)(1,0.2332281684)(2,0.07540113128)(3,0.05992425693)(4,0.0402675323)(5,0.0266118958)(6,0.0260453425)(7,0.01221493768)(8,0.01220673116)(9,0.008715620032)(10,0.008289185336)(11,0.007870957153)(12,0.00754330455)(13,0.00461479543)(14,0.004067390663)(15,0.004063135435)(16,0.003931223348)(17,0.003554027726)(18,0.002907232977)(19,0.002346454635)(20,0.002197217688)(21,0.001946463145)(22,0.001934609294)(23,0.001619114487)(24,0.001615163203)(25,0.00154920716)(26,0.001276264639)(27,0.001082955682)(28,0.001065326878)(29,0.0009945077156)(30,0.00091973727)(31,0.0008595561797)(32,0.0007884330729)(33,0.0007583425277)(34,0.0006905628148)(35,0.0005194418356)(36,0.0005127550478)(37,0.0004656435881)(38,0.0004562212962)(39,0.0004188360734)(40,0.0004176202938)(41,0.0003695969995)(42,0.0002993857274)(43,0.0002869239864)(44,0.0002583531657)(45,0.0002580492208)(46,0.0001936129019)(47,0.0001914852875)(48,0.0001875340038)(49,0.0001659539159)(50,0.0001525803402)(51,0.0001474132769)(52,0.0001100280541)(53,0.00008935980086)(54,0.00008935980086)(55,0.00008540851714)(56,0.00004923907394)
};
\addlegendentry{$N=\infty$}

\end{axis}
\end{tikzpicture}
    \caption{Class probabilities $p(k)$ sorted from highest to lowest for all 57 
    classes as $N$ in Equation~\ref{eq:2} is varied from 
    $\infty$ (original distribution) to 1,000.}
    \label{fig:dis}
\end{figure}
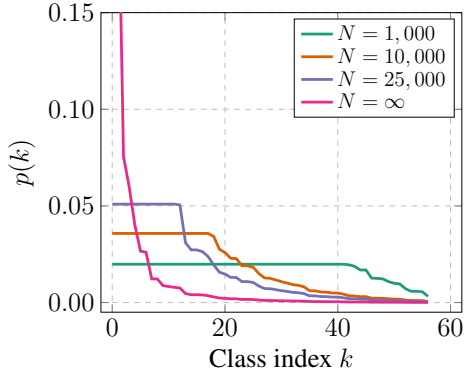

{\bf Class Imbalance.} The object detection dataset contains significant 
class imbalance with a long tail. Randomly sampled mini-batch training with 
skewed class distribution over emphasizes frequent classes. Since 
challenge objective assigns equal weight to every class, model bias towards
popular classes can significantly hurt performance. To address this problem we adopt a 
sampling strategy that approximately balances class distribution during training.
Let $n_k$ denote the number of images containing class $k$ in the training set 
with $K$ classes. Randomly sampling from the training set results in the
following probability for each class:
\begin{equation}
  p(k) = \frac{n_k}{\sum_{i=1}^{K }n_i} \label{eq:1}
\end{equation}
Frequent classes with high image count $n_k$ thus have much higher probability 
of being included in each mini-batch. To balance the probabilities we 
introduce an additional parameter $N$ and sample images according to:
\begin{equation}
  p(k) = \frac{\min(n_k, N)}{\sum_{i=1}^{K}\min(n_i, N)}~\label{eq:2}
\end{equation}
The effect of $N$ is shown in Figure~\ref{fig:dis}. The figure shows class
probabilities for all 57 classes sorted from highest to lowest as
$N$ is varied from $\infty$ (original distribution) to 1,000.
We see a gradual effect where class distribution approaches uniform
distribution as $N$ is decreased. Empirically, we found that 
setting $N$ in the [1,000, 10,000] range produced better performance
than using original or uniform distribution.
Compared to the original class distribution, the Cascade RCNN model
improved in performance from $0.44$ to $0.53$ on the validation set, 
and from $0.058$ to $0.065$ on the object detection leaderboard.

{\bf Learning Setup.} We split the challenge dataset into 374,768
triplets to train the second stage spatio-semantic and visual models, and 12,314
triplets to train the third stage aggregation model. All models are validated 
on the remaining 3,991 triplets. To evaluate model performance we 
use the competition 
metric~\footnote{\url{https://storage.googleapis.com/openimages/web/evaluation.html}}
which aims to capture both object and relationship detection quality. 
In all experiments we use the $\text{mAP}_\text{rel}$ component of the
metric to validate all models. We found it to correlate well with the 
overall metric and much faster to compute.

\subsection{Implementation Details}\label{sec:impl}

Our pipeline consists of three stages that include object detection,
spatio-semantic and visual information extraction, and final aggregation.
In this section we describe the implementation details for each stage.

{\bf Object Detection.} For the object detection stage, we use an 
ensemble of Cascade RCNN~\cite{SunXLW19} detection networks with 
ResNeXt \cite{Xie2016} and HRNet~\cite{WangSCJDZLMTWLX19} backbones 
trained on COCO and fine-tuned using our partial weight transfer
approach. As described in Section~\ref{sc:pwt}, we are able to
initialise 44 out of the 57 challenge classes by mapping them onto
the 80 COCO classes. The other 13 classes are either initialized 
randomly or transferred from the backbone that is fine-tuned 
for the challenge classes without partial weight transfer.
To combine multiple detection models we use a weighted non-maximum suppression (NMS) 
approach where bounding boxes from all detection models are combined using a weighted average.
The weights for each model are selected according to performance on the 
validation set. This approach is similar to traditional NMS except instead 
of choosing the most confident box we use weighted average. Empirically, we found
that weighted average provided a gain of around 2 points on the leaderboard.
\begin{table}
\centering
\begin{tabular}{llcc}
\hline
Rank & Team & Public LB & Private LB \\
\hline
1& Layer6 AI & {\bf 0.4638} & {\bf 0.4080}\\
2& tito & 0.4407 & 0.3881\\
3& Very Random team & 0.4289 & 0.3785\\
4& [ods.ai] n01z3 & 0.3984 & 0.3659\\
5& Ode to the Goose & 0.4016 & 0.3477\\
\hline
\end{tabular}
\vskip 0.1cm
\caption{Final team rankings on the public and private leaderboards.}
\label{table:lb}
\end{table}

For training of visual relationship models, we use ground truth bounding
boxes instead of predicted ones. We also experiment with using the predicted
boxes which we expect to perform better. The rationale is that exposing the 
relationship model to errors (e.g. shifted bounding boxes or mislabeled classes) 
made by object detection should enable it to learn to correct them 
and make more robust predictions. However, this does not perform well,
and we presume that this is because the relationship model is not able
to sufficiently correct errors made earlier in the pipeline.

{\bf Special ``is'' Relationship.} We use a separate pipeline for the 
``is" relationship since, unlike other relationships, it doesn't operate 
on pairs of objects. We leverage the object detection model and modify the
classification head to predict over all object-attribute pairs that can form
a valid ``is" relationship as separate classes. One concern here is that 
there aren't enough training examples to learn a reliable 
detection model for each pair. We address this problem by again leveraging the 
partial weight transfer approach. This time we transfer weights from one of our base 
detection models and then fine-tune on the available ``is" relationship 
training data. For instance, both "wooden" and "plastic" piano classes get 
initialized with the piano classifier weights from our base detection model 
as well as its backbone. Fine-tuning this way makes the model more robust to
lack of training data and improves performance.

{\bf Spatio-Semantic and Visual Models.}
For spatio-semantic model we use the tree-based GBM architecture from the 
XGBoost library~\cite{Chen:2016:XST:2939672.2939785} due to its excellent 
performance in our experiments. We train a separate model for each 
relationship type by framing the problem into binary classification.
Specifically, we iterate over all ground truth object bounding boxes 
and for each pair of objects that can form the target relationship we 
check whether that pair is in the ground truth training relationship set.
If it is, we label it as a positive sample, and if it is not as a 
negative sample. Note that negative samples are approximate 
here since two objects can have a relationship that is not labelled 
in the training set. However, the probability of that is small and 
empirically we found that using this procedure with negative samples 
produced good performance. We use the same set of hyper-parameters 
for all relationships, the GBM 
model is trained with the \texttt{dart} booster and \texttt{max\_depth} 
set to $10$. To prevent over-fitting, we further set \texttt{subsample} 
and \texttt{colsample\_bytree} parameters to $0.2$, as well as 
\texttt{gamma} and \texttt{lambda} parameters to $2.0$ and $1000$ 
respectively. Each model is trained for $5000$ boosting iterations with 
an early stopping check every $50$ iterations.
\begin{table}
\centering
\begin{tabular}{lcccc}
\hline
\multirow{2}{*}{Relationship} &Spatio- & \multirow{2}{*}{Visual} &\multirow{2}{*}{Avg.} &\multirow{2}{*}{3'rd Stage} \\
& Semantic & & & \\
\hline
plays & 0.49 & 0.58 & 0.55 & \textbf{0.59} \\
hits & 0.58 & 0.47 & 0.58 & \textbf{0.61} \\
at & 0.37 & 0.35 & 0.35 & \textbf{0.42} \\
inside\_of & 0.31 & 0.35 & 0.32 & \textbf{0.37} \\
interacts\_with & 0.42 & 0.42 & 0.41 & \textbf{0.44} \\
\hline
\end{tabular}
\vskip 0.1cm
\caption{Validation $\text{AP}_\text{rel}$ results 
for a subset of five relationships. We show performance for spatio-semantic and 
visual models, and two ways of combining them using weighted average (Avg.) 
and 3'rd stage GBM model.}
\label{table:vis}
\end{table}

For the visual model we use the ResNeXt backbone from the object detection 
model that has been fine-tuned for the challenge classes. We apply a 3-layer 
MLP on top of the backbone with ReLU activations. The last layer outputs a 
binary sigmoid prediction, and we train this model using the same 
positive/negative samples as the spatio-semantic model. For all relationships, we use the same 
batch size of $32$ and run optimization for $35$ epochs. To reduce 
overfitting, we apply dropout with $p = 0.2$ to each MLP layer. We use the 
Adam optimizer with cosine learning rate annealing and linear learning 
rate warmup. The maximum and the minimum learning rates are 
$3\mathrm{e}^{-4}$ and $5\mathrm{e}^{-5}$ respectively.

{\bf Final Aggregation.} In the last stage we combine predictions 
from spatio-semantic and visual models together with object features 
(see Section~\ref{sc:spatio-semantic}) to generate the final
relationship prediction. We also use a tree-based GBM model here, 
and train it with the XGBoost library. The parameters for this model 
are the same as for the spatio-semantic model, with the only difference
that we use \texttt{gbtree} booster instead of \texttt{dart} and 
lower tree depth to $8$.

\subsection{Results}\label{sec:results}

The final team standings are shown in Table~\ref{table:lb}. Our team 
``Layer6 AI" outperforms all other teams on both public and private leaderboards 
beating the second place team by over 5\%. These results indicate
that our multi-stage pipeline is highly robust and produces leading 
performance on this challenging task. We can also conclude that by applying 
transfer learning through our partial weight transfer approach we can
train highly accurate visual relationship models with minimal hardware requirements.
\begin{table}
\centering
\begin{tabular}{lclclc}
\hline
\multirow{2}{*}{Attribute} & \multirow{2}{*}{$\text{AP}_\text{rel}$} & \multirow{1}{*}{Best} &\multirow{1}{*}{Best} &\multirow{1}{*}{Worst} &\multirow{1}{*}{Worst} \\
& & Class & $\text{AP}_\text{rel}$ & Class & $\text{AP}_\text{rel}$ \\
\hline
transparent & 0.40 & bottle & 0.75  & table & 0.13 \\
wooden & 0.60 & guitar & 0.95 & bench & 0.10 \\
plastic & 0.39 & piano & 0.68 & bench & 0.02 \\
leather & 0.46 & sofa & 0.68 & suitcase & 0.25 \\
textile & 0.62 & sofa & 0.80 & suitcase & 0.40 \\
\hline
\end{tabular}
\vskip 0.1cm
\caption{Breakdown of the "is" model results by attribute. For each 
attribute we show validation $\text{AP}_\text{rel}$ results together 
with best and worst performing class.
}
\label{table:is}
\end{table}

\begin{figure*}[t]\centering
\subfloat[t][\label{fig:qual_1}]{%
           \begin{tikzpicture}[baseline]
	\begin{scope}[node distance=1cm and 0.1cm]
	\definecolor{c_subject}{rgb}{0.76,0.93,0.78}
	\definecolor{c_object}{rgb}{0.79,0.62,0.92}
	\definecolor{c_relation}{rgb}{0.38,0.83,0.90}
\tikzstyle{relation}=[rectangle,draw=none,fill=c_relation,opacity=0.7,text opacity=1, anchor=south west,inner sep=0.7pt]
\tikzstyle{subject}=[rectangle,draw=none,fill=c_subject, opacity=0.7,text opacity=1, anchor=south west,inner sep=0.7pt]
\tikzstyle{object}=[rectangle,draw=none,fill=c_object, opacity=0.7,text opacity=1, anchor=south west,inner sep=0.7pt]
            \node (a) {\includegraphics[height=4.7cm]{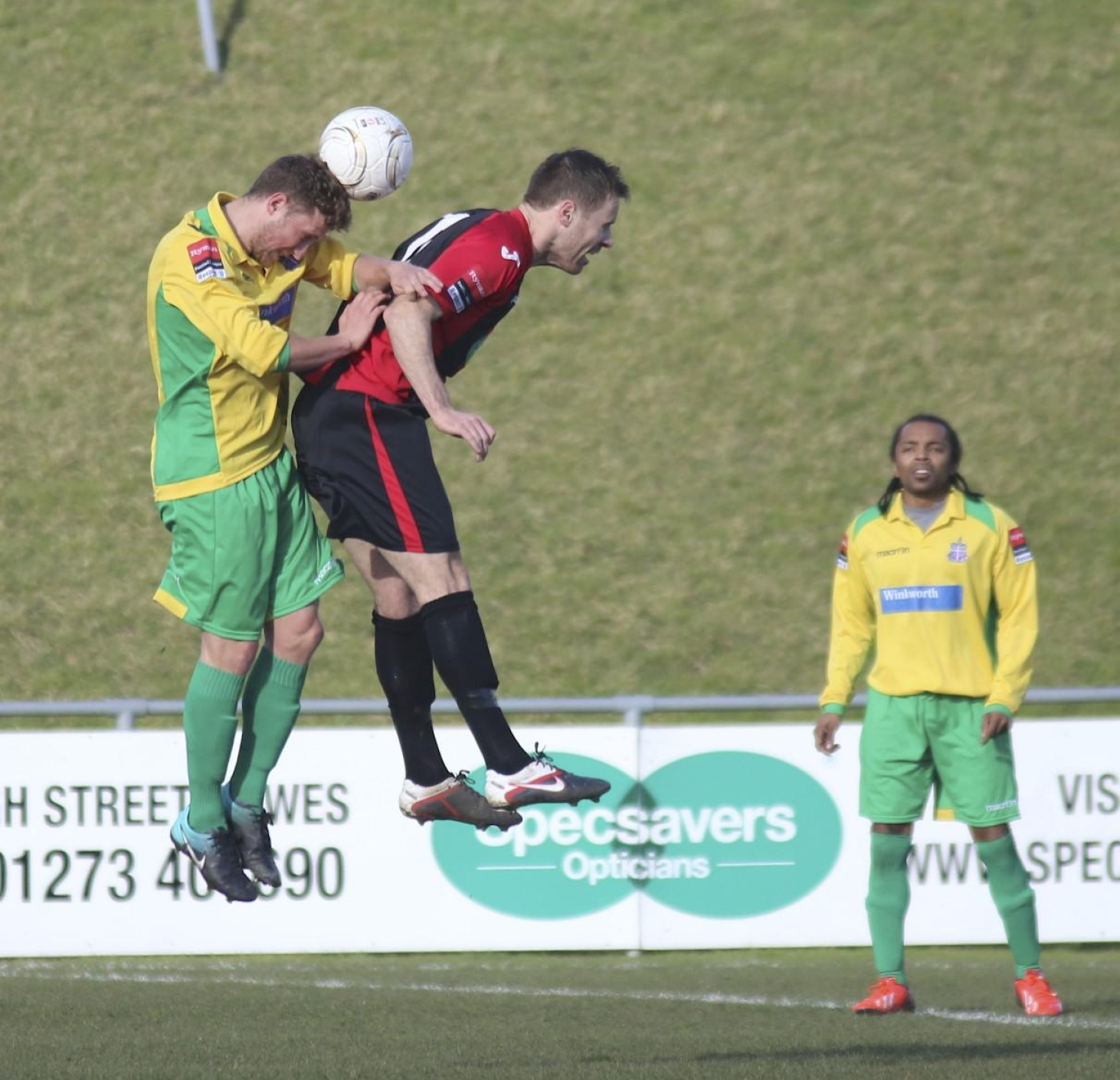}};

        \draw[c_relation,ultra thick,opacity=0.7] ($ (a) + (-1.875,-1.59) $) rectangle ($ (a) + (-0.545,1.95) $);
        \draw[c_subject,ultra thick,opacity=0.7] ($ (a) + (-1.855,-1.555) $) rectangle ($ (a) + (-0.585,1.675)$);
        \draw[c_object,ultra thick,opacity=0.7] ($ (a) + (-1.11,1.445) $) rectangle ($ (a) + (-0.67,1.915)$);

        \node at ($ (a) + (-1.875,1.95) $) [relation] {\scriptsize Hits};
        \node at ($ (a) + (-1.855,1.675) $) [subject] {\scriptsize Man};
        \node at ($ (a) + (-1.11,1.875) $) [object] {\scriptsize Football};

            \end{scope}
        \end{tikzpicture}
}
\hfill
\subfloat[t][\label{fig:qual_2}]{%
    \centering
           \begin{tikzpicture}[baseline]
	\begin{scope}[node distance=1cm and 0.1cm]
	\definecolor{c_subject}{rgb}{0.76,0.93,0.78}
	\definecolor{c_object}{rgb}{0.79,0.62,0.92}
	\definecolor{c_relation}{rgb}{0.38,0.83,0.90}
\tikzstyle{relation}=[rectangle,draw=none,fill=c_relation,opacity=0.7,text opacity=1, anchor=south west,inner sep=0.7pt]
\tikzstyle{subject}=[rectangle,draw=none,fill=c_subject, opacity=0.7,text opacity=1, anchor=south west,inner sep=0.7pt]
\tikzstyle{object}=[rectangle,draw=none,fill=c_object, opacity=0.7,text opacity=1, anchor=south west,inner sep=0.7pt]
        
         \node (b) {\includegraphics[height=4.7cm]{1_raw.png}};
\draw[c_relation,ultra thick, opacity=0.7]($(b) + (-3.705,-1.65) $) rectangle ($ (b) + (-0.675,2.05) $);
\draw[c_subject,ultra thick, opacity=0.7]($(b) + (-3.655,-1.565) $) rectangle ($ (b) + (-1.305,1.145) $);
\draw[c_relation,ultra thick, opacity=0.7]($(b) + (-1.76,0.345) $) rectangle ($ (b) + (-0.1,2.055) $);
\draw[c_object,ultra thick, opacity=0.7]($(b) + (-1.7,0.945) $) rectangle ($ (b) + (-0.72,1.955) $);
\draw[c_subject,ultra thick, opacity=0.7]($(b) + (-1.225,0.385) $) rectangle ($ (b) + (-0.175,1.195) $);
        \node at ($ (b) + (-3.705,2.05) $) [relation] {\scriptsize Hits};
        \node at ($ (b) + (-1.76,2.055) $) [relation] {\scriptsize Hits};
        \node at ($ (b) + (-3.655,1.145) $) [subject] {\scriptsize Man};
        \node at ($ (b) + (-1.225,1.195) $) [subject] {\scriptsize Man};
        \node at ($ (b) + (-1.7,1.7) $) [object] {\scriptsize Football};
            \end{scope}
        \end{tikzpicture}
}
\hfill
\subfloat[t][\label{fig:qual_3}]{%
           \begin{tikzpicture}[baseline]
	\begin{scope}[node distance=1cm and 0.1cm]
	\definecolor{c_subject}{rgb}{0.76,0.93,0.78}
	\definecolor{c_object}{rgb}{0.79,0.62,0.92}
	\definecolor{c_relation}{rgb}{0.38,0.83,0.90}
\tikzstyle{relation}=[rectangle,draw=none,fill=c_relation,opacity=0.7,text opacity=1, anchor=south west,inner sep=0.7pt]
\tikzstyle{subject}=[rectangle,draw=none,fill=c_subject, opacity=0.7,text opacity=1, anchor=south west,inner sep=0.7pt]
\tikzstyle{object}=[rectangle,draw=none,fill=c_object, opacity=0.7,text opacity=1, anchor=south west,inner sep=0.7pt]
            \node (c) {\includegraphics[height=4.7cm]{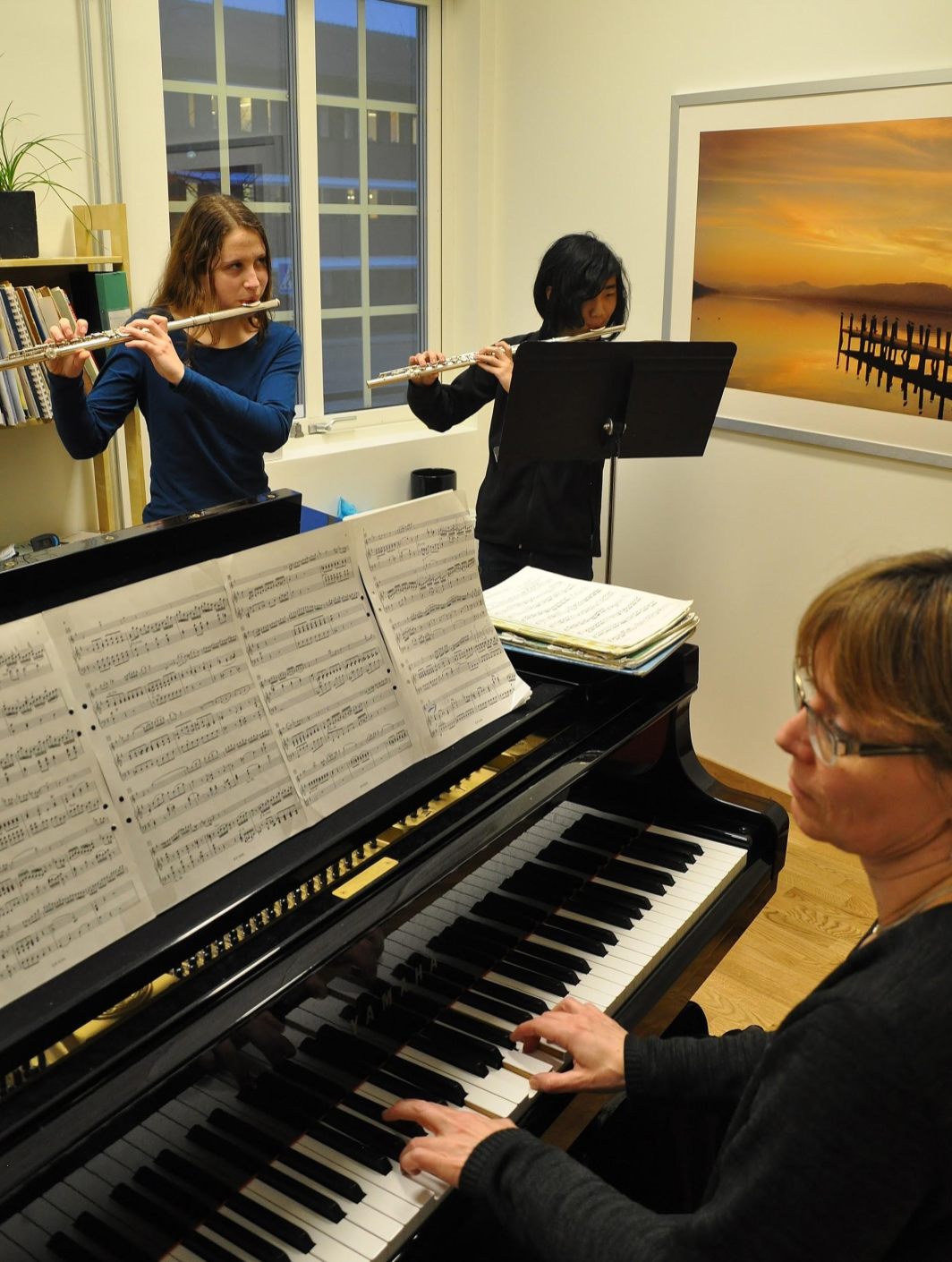}};
\draw[c_subject,ultra thick, opacity=0.7]($(c)+(-1.745,0.425) $) rectangle ($(c)+(-0.635,1.615) $);
\draw[c_relation,ultra thick, opacity=0.7]($(c)+(-1.765,0.375) $) rectangle ($(c)+(-0.595,1.665) $);
\draw[c_object,ultra thick, opacity=0.7]($(c)+(-1.745,0.95) $) rectangle ($(c)+(-0.735,1.27) $);
\draw[c_subject,ultra thick, opacity=0.7]($(c)+(-0.415,0.195) $) rectangle ($(c)+(0.595,1.485) $);
\draw[c_relation,ultra thick, opacity=0.7]($(c)+(-0.475,0.12) $) rectangle ($(c)+(0.635,1.54) $);
\draw[c_object,ultra thick, opacity=0.7]($(c)+(-0.425,0.89) $) rectangle ($(c)+(0.585,1.15) $);
\draw[c_subject,ultra thick, opacity=0.7]($(c)+(-0.355,-2.295) $) rectangle ($(c)+(1.695,0.335) $);
\draw[c_relation,ultra thick, opacity=0.7]($(c)+(-1.78,-2.355) $) rectangle ($(c)+(1.74,0.535) $);
\draw[c_object,ultra thick, opacity=0.7]($(c)+(-1.775,-2.295) $) rectangle ($(c)+(1.175,0.495) $);
\node at ($(c)+(-1.745,0.95)$)[object,anchor=north west]{\scriptsize Flute};
\node at ($(c)+(-1.745,1.42)$)[subject]{\scriptsize Woman};
\node at ($(c)+(-1.765,1.65)$)[relation]{\scriptsize Holds};

\node at ($(c)+(0.595,1.32)$)[subject]{\scriptsize Woman};
\node at ($(c)+(0.635,1.55)$)[relation,anchor=south east]{\scriptsize Holds};
\node at ($(c)+(0.585,1.1)$)[object,anchor=west]{\scriptsize Flute};

\node at ($(c)+(1.695,0.335)$)[subject,anchor=south east]{\scriptsize Woman};
\node at ($(c)+(1.74,0.535)$)[relation,anchor=south east]{\scriptsize Plays};
\node at ($(c)+(1.175,0.1)$)[object,anchor=south east]{\scriptsize Piano};
        
            \end{scope}
        \end{tikzpicture}
}
\caption{
Qualitative results from our model. Green and purple bounding boxes are 
detected subject and object classes, and purple box is the predicted 
relationship between them.
}\label{fig:qual}
\end{figure*}
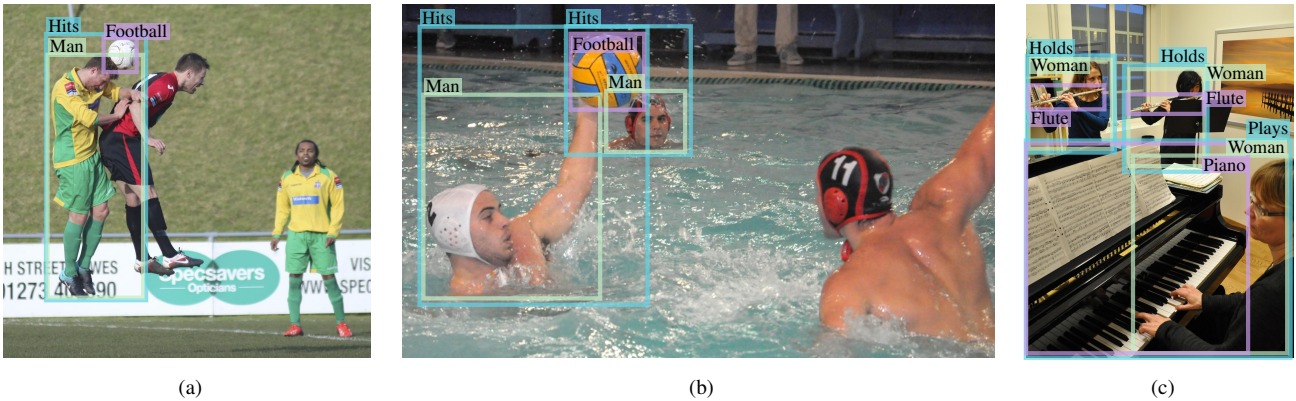

Table~\ref{table:vis} summarizes performance for each stage across 
a sample of five relationships. We see that the spatio-semantic model performs 
better on ``at'' and ``hits'' relationships, while visual model outperforms 
on ``plays'' and ``inside\_of''. This validates our hypothesis that both 
types of information are required to accurately detect all relationships. We 
also see that combining these models through averaging is highly sub-optimal 
and actually hurts performance for four out of the five relationships. This 
observation motivates us to introduce a third stage to learn how 
to better combine spatio-semantic and visual predictions. Results for the
third stage are shown on the right in Table~\ref{table:vis}, we see that the 
third stage model is able to effectively learn when to use each type of signal and 
further improve performance. We are able to consistently improve performance over the best 
individual model on all five relationships, with particularly significant 
improvement on the ``at" relationship where we gain over 5 points in
$\text{mAP}_\text{rel}$ or 11\%.

We described in Section~\ref{sec:impl} that the ``is" relationship is treated differently 
in our pipeline since instead of pairs it operates on object-attribute combinations. 
Table~\ref{table:is} shows the $\text{AP}_\text{rel}$ performance of the ``is" model for each 
of the five attributes. For each attribute we also show best and worst object class with 
corresponding $\text{AP}_\text{rel}$. From the table we see that
that the model performs well on attributes wooden and textile, and does significantly worse 
on transparent and plastic. As expected performance is highly dependent on the number of training
instances for each object-attribute pair, as well as label ambiguity. For instance, many plastic objects
such as bottles are also labelled as transparent, and the model has difficulty distinguishing
between the two properties. By analysing the best and worst class for each attribute we can directly
observe the effect of the training set size. Textile sofas and suitcase appear much more frequently in
the training data (and arguably in real life) than leather ones so performance on textile is better. 
Interestingly, the model has difficulty recognizing attributes such as transparent, wooden and 
plastic for common furniture items such as tables and benches. After inspecting the data we observed
that furniture objects have a lot of variably, and often appear in cluttered scenes with many occlusions 
making it challenging to identify what they are made off.

Qualitative examples are shown in Figure~\ref{fig:qual}. Figure~\ref{fig:qual}a shows a difficult  
scene where two soccer players are trying to get the ball but only one of them hits it. 
Our model is able to correctly identify which player hit the ball which indicates a degree 
of robustness to spatially complex scenes.
Figure~\ref{fig:qual}b shows a related failure case where player hits the ball during a game
of water polo. The model is able to correctly capture that relationships, but also identifies another
player as hitting the same ball which is incorrect. Possible reasons for this failure can be 
partial occlusion between the two players, and position of the incorrectly identified player
relative to the ball. Position in particular is difficult to capture accurately here, the two
bounding boxes are close together in 2D but the player is actually far from the ball in 3D. 
3D spatial information is difficult to capture with a single image and we hypothesise
that performance can be improved if another view or depth information is added as input.
Finally, Figure~\ref{fig:qual}c shows a more cluttered scene where multiple musicians are 
playing various instruments such as flute and piano. Here, we see that our model is able to 
correctly identify all relationships even though musicians are in close proximity to
each other.

\section{Conclusion}

We present our winning solution to the Open Images 2019 Visual Relationship challenge.
We propose a novel partial weight transfer approach to effectively transfer learned
models between datasets and accelerate training. Our pipeline consists 
of object detection followed by spatio-semantic and visual feature extraction, 
and a final aggregation phase where all information is combined to generate 
relationship prediction. Partial weight transfer enables us to train the 
entire architecture in under two days on a {\it single} GPU making it accessible to
most researchers and practitioners. In addition to high efficiency, 
we also achieve top performance and beat over 200 teams to place
first in the competition outperforming the second place team by over 5\%. In the
future work we aim to focus on fusing the three stages into a joint architecture
that can be trained end-to-end. We hypothesize that end-to-end training can improve
the flow of information and lead to better performance.

\newpage

{\small
\bibliographystyle{ieee_fullname}
\bibliography{challenge_v2}
}

\end{document}